\def\year{2021}\relax
\documentclass[letterpaper]{article} 
\usepackage{aaai21}  
\usepackage{times}  
\usepackage{helvet} 
\usepackage{courier}  
\usepackage[hyphens]{url}  
\usepackage{graphicx} 
\usepackage[colorinlistoftodos]{todonotes}
\usepackage{natbib}  
\usepackage{caption} 
\usepackage{amsmath}
\usepackage{amssymb}
\usepackage{amsthm}
\usepackage{algorithm}
\usepackage[noend]{algpseudocode}
\usepackage{subcaption}

\DeclareMathOperator*{\argmax}{arg\,max}
\DeclareMathOperator*{\argmin}{arg\,min}

\newtheorem{theorem}{Theorem}

\newtheorem{lemma}[theorem]{Lemma}
\newtheorem{definition}[theorem]{Definition}
\newtheorem{assumption}[theorem]{Assumption}
\usepackage{booktabs}
\usepackage{xr}
\makeatletter
\newcommand*{\addFileDependency}[1]{
  \typeout{(#1)}
  \@addtofilelist{#1}
  \IfFileExists{#1}{}{\typeout{No file #1.}}
}
\makeatother

\newcommand*{\myexternaldocument}[1]{
    \externaldocument{#1}
    \addFileDependency{#1.tex}
    \addFileDependency{#1.aux}
}

\myexternaldocument{appendix}

\title{Theoretically Principled Deep RL Acceleration via\\ Nearest Neighbor Function Approximation}
\author {
    Junhong Shen, Lin F. Yang\\
}
\affiliations {
    University of California, Los Angeles\\
    jhshen@ucla.edu, linyang@ee.ucla.edu
}

\begin{document}
\maketitle
\begin{abstract}
Recently, deep reinforcement learning (RL) has achieved remarkable empirical success by integrating deep neural networks into RL frameworks. However, these algorithms often require a large number of training samples and admit little theoretical understanding. To mitigate these issues, we propose a theoretically principled nearest neighbor (NN) function approximator that can improve the value networks in deep RL methods. Inspired by human similarity judgments, the NN approximator estimates the action values using rollouts on past observations and 
can provably obtain a small regret bound that depends only on the intrinsic complexity of the environment. We present (1) Nearest Neighbor Actor-Critic (NNAC), an online policy gradient algorithm that demonstrates the practicality of combining function approximation with deep RL, and (2) a plug-and-play NN update module that aids the training of existing deep RL methods. Experiments on classical control and MuJoCo locomotion tasks show that the NN-accelerated agents achieve higher sample efficiency and stability than the baseline agents. Based on its theoretical benefits, we believe that the NN approximator can be further applied to other complex domains to speed-up learning.
\end{abstract}

\section{Introduction}
People learn a variety of relationships in life, e.g., we associate the force of pressing the gas pedal with the amount of acceleration gained while driving. In the context of reinforcement learning (RL), where an agent interacts with the environment to maximize the cumulative rewards, the learning objective is the relationship between state-action pairs and future gains. Theories on associative learning suggest that people learn from similarity measures \cite{carroll1963functionlearning, busemeyerbdm97}: if $x$ can predict $y$, they presume that observations similar to $x$ have similar $y$ values. It is thus natural to consider integrating similarity-based models into active learning. A suitable choice for the RL setting is the nearest neighbor function approximator \cite{Emigh2016ReinforcementLI, Shah2018QlearningWN, yang2019learning}. 

We study online episodic RL with unknown reward and transition functions. Existing deep RL algorithms achieve impressive results in robot control \citep[e.g.,][]{Lillicrap2016ContinuousCW, levine2016robotics, Gu2017DeepRL}, Go \cite{Silver2016MasteringTG} and Atari playing \cite{Mnih2013PlayingAW}. However, several challenges still exist. First, the theoretical foundation of deep RL has not been fully established
\cite{Arulkumaran2017ABS, Lake2018BuildingMT}. It is often mysterious why an algorithm works or fails in certain cases \cite{Kansky2017SchemaNZ}. Second, empirical results suggest that model-free deep RL often requires considerable samples to learn \cite{Deisenroth2011PILCOAM}. Nonetheless, data can be expensive to acquire in practical domains like healthcare \cite{Kober2013ReinforcementLI}. High-dimensional input, such as pixel data, demands a larger sample size even if the problem itself is simple \cite{Lillicrap2016ContinuousCW}. Third, online learning coupled with neural networks is generally regarded as unstable \cite{Hasselt2016DeepRL}. Hyperparameter tuning can also affect the learning outcome.
In sum, improving deep RL with theory-based approaches is of critical importance.

On the other hand, great theoretical progress has been made in tabular RL, where the state-action spaces are small and finite. The sample complexity of tabular methods are properly studied \citep[e.g.,][]{Jaksch2008NearoptimalRB, Azar2017MinimaxRB, Jin2018IsQP,Zanette2019TighterPR}. Yet the best obtainable complexity depends linearly on the number of states, which can be huge in reality, e.g., the game of Go has about $3^{19\times19}$ states. Thus, real-world application of tabular theories remains a challenge.

In this paper, we bridge the gap between RL theory and practice by a theoretically principled deep RL acceleration technique. Specifically, we exploit the structural properties of the state-action space by using the nearest neighbor (NN)\footnote{Throughout the paper, the abbreviation ``NN" is used to refer to ``nearest neighbor," not ``neural network."} approximator for value estimation. Such a function approximator not only attains a theoretical guarantee in sample complexity but also possesses good generalization ability when plugged into deep RL frameworks. 
In fact, we show that the NN approximator with an upper-confidence construction obtains a near-optimal regret $O[H(DLK)^{d/(d+1)}]$ for both low- and high-dimensional inputs in deterministic systems, where $K$ is the number of episodes, $H$ is the episode length, $L$ is a Lipschitz constant related to the distance metric measuring state similarities, $D$ and $d$ are respectively the diameter and dimension of the intrinsic state-action space. 

We demonstrate the empirical efficacy of the NN approximator by fitting it into the actor-critic framework. We use the non-parametric NN critic to bootstrap state values and train the policy network with temporal-difference methods \cite{Sutton1988LearningTP}. Given a state-action pair, the NN critic finds within history the closest sample to this observation under the distance metric. The corresponding reward plus a Lipschitz confidence term and the next state are used to approximate the reward and transition functions, respectively. The algorithm displays impressive learning speed in the cart-pole balancing problem. Beyond this, we also encapsulate the NN approximator into a plug-and-play module that boosts the training of existing deep RL algorithms without changing their original structures. The plug-in NN critic encourages action exploration and stabilizes value learning. We evaluate the module with state-of-the-art deep RL agents on a set of 3D locomotion tasks. Results show that the NN-aided gradient update improves both training speed and stability.

\subsubsection{Roadmap:} The paper is organized as follows: we first discuss related works in theoretical and deep RL. Next, we introduce notations and concepts in RL and metric dimensions. We then give the formulation of the NN approximator and analyze its theoretical guarantee. In the later sections, we present two algorithms that combine the NN approximator with deep RL and evaluate their empirical performance.

\section{Related Work}
\subsubsection{RL with neural networks:}
There is a long line of research that applies deep RL to games and control problems \citep[e.g.,][]{ Mnih2013PlayingAW, schulman2015TRPO, levine2016robotics, Hasselt2016DeepRL, Lillicrap2016ContinuousCW}.
These results employ several heuristics to accelerate exploration. For example, \citet{Mnih2013PlayingAW, Mnih2015HumanlevelCT, Hasselt2016DeepRL} randomly sample actions and store them in a replay buffer before policy learning. Gaussian noise \cite{Lillicrap2016ContinuousCW} or Ornstein-Uhlenbeck noise \cite{Fujimoto2018AddressingFA} are added to the actions or the network parameters \cite{Plappert2018ParameterSN} to encourage exploration. Several works also combine model-based value estimation and model-free policy learning to reduce sample complexity \cite{Deisenroth2011PILCOAM, Buckman2018SampleEfficientRL}. 

Although there is limited theoretical understanding of the aforementioned methods, we emphasize that our aim is not to improve them but rather introduce a new theory-based function approximator to the literature of deep RL. By combining NN value estimation and existing frameworks, we believe that the efficiency of model-free RL algorithms can be improved in a provable manner (at least in some settings).

\subsubsection{RL with theoretical guarantees:} 
To facilitate theoretical analysis, many works study RL in the tabular setting, where the state and action spaces are discrete \citep[e.g.,][]{Jaksch2008NearoptimalRB, Azar2017MinimaxRB, Jin2018IsQP,Zanette2019TighterPR}. 
The sample complexity of the algorithms depends at least linearly on the number of states ${|\mathcal{S}|}$. Since this number tends to be large in practice, it is difficult to extend these algorithms to real-world settings.
Recently, several works have emphasized understanding in RL with general function approximation \citep[e.g.,][]{Osband2014ModelbasedRL, Jiang2017ContextualDP, Sun2019ModelbasedRI, Wang2020OnRR}. However, the function classes are either simple, e.g., linear functions \citep[e.g.,][]{Yang2019SampleOptimalPQ, Jin2020ProvablyER}, or have strong structural assumptions, which prevent them from being applied to more practical problems.

\subsubsection{RL with nearest neighbor search:}
Combining nearest neighbor search with active learning has been studied in episodic RL. Model-free episodic control \cite{Blundell2016ModelFreeEC} builds on a tabular memory and applies regression using the mean of k-nearest neighbors for Q-value estimation. 
Neural episodic control \cite{Pritzel2017NeuralEC} and episodic memory deep Q-networks \cite{Lin2018EpisodicMD} improve the algorithm's generalization ability by absorbing state features into networks. These algorithms take the NN search as a pure classification technique. They do not exploit the fact that the distance between state-action pairs can indicate their relative values. In addition, the value estimation for these methods exists at the trajectory level: the Q-values are updated at the end of an episode by the total reward of a trajectory. In contrast, the NN value estimation in this paper takes the form of on-policy Monte Carlo rollouts using samples from independent environment steps. Thus, we leverage not only intra-episode but also inter-episode information. 

The prototype of our NN function approximator is presented in \citet{yang2019learning}, where an upper-confidence algorithm with general function approximation is proposed for tabular RL. The algorithm can apply to continuous cases but requires a discretization of the action space. We improve it to account for non-tabular cases without discretizing the action space. Meanwhile, though \citet{yang2019learning} derive a regret based on the ambient dimension of the state space, they do not justify the regret of high-dimensional inputs with small intrinsic dimensions. 

\section{Preliminaries}
In this section, we introduce the key definitions and notations in RL. For the clarity of the proofs, we assume that the Markov decision process (MDP) is finite-horizon and deterministic. This assumption is not restrictive, as many real-world control systems do not involve randomness. 

Formally, we consider an MDP $(\mathcal{S}, \mathcal{A}, f, r, H)$ with state space $\mathcal{S}$, action space $\mathcal{A}$, deterministic transition model $f: \mathcal{S} \times \mathcal{A} \rightarrow \mathcal{S}$, and reward function $r: \mathcal{S} \times \mathcal{A} \rightarrow \mathbb{R}$. An agent interacts with the environment episodically, where each episode lasts $H$ steps. 
In an episode, the agent starts from an initial state $s_1$ independent of the history. At step $h \in [H]$\footnote{$[H]$ denotes the set of integers $\{1, ... , H\}$.}, it observes state $s_h$ and performs action $a_h:=\pi(s_h, h)$ according to the policy $\pi: \mathcal{S}\times[H] \rightarrow \mathcal{A}$. It then receives reward $r_h = r(s_h, a_h)$ and next state $s_{h+1} = f(s_h, a_h)$. We define the cumulative reward from $h$ as $R_h = \sum_{t = 0}^{H-h} r_{h+t}$. The goal of learning is to find a policy that maximizes the total reward in one episode when $f$ and $r$ are unknown.

Given a policy $\pi$, the value function $V^\pi: \mathcal{S} \times [H] \rightarrow \mathbb{R} $ is defined as the cumulative reward from state $s$ at step $h$ and following $\pi$ therefrom. It satisfies the Bellman equation:
\begin{align}
    V^\pi_h(s) = r(s, \pi(s)) + V^\pi_{h+1}[f(s, \pi(s))],
    \label{eq:BellmanV}
\end{align}
with $V_H^\pi(s) = r(s, \pi(s))$. The optimal policy $\pi^*$ is the one such that $V^{\pi^*}_h := V^*_h = \max_{\pi} V^\pi_h$. The temporal-difference (TD) error at step $h$ is $\delta_h = r_h + V^\pi_{h+1}(s_{h+1}) - V^\pi_h(s_h)$.

We further denote the action value (or Q-function) as $Q^\pi_h(s, a) := r(s, a) + V^\pi_{h+1}\big(f(s, a)\big)$. The optimal $Q^*_h := \max_\pi  Q^\pi_h$ gives the maximum values for a $(s,a)$ pair achievable by any policy. By the Bellman optimality equation, we have $V^*_h(s)= \max_{a \in A}Q^*_h(s,a)$.

We measure the sample complexity of an algorithm by its regret, which is the difference between the total rewards of the unknown optimal policy and that gathered in learning:
\begin{align*} 
    \mathrm{Regret}(K) = \sum_{k=1}^K [V^*_1(s^k_1) - \sum_{h=1}^H r(s^k_h, a^k_h)],
\end{align*}
where $K$ is the number of episodes played. 

\section{Theoretical Guarantee of Nearest Neighbor Function Approximation}
In this section, we first introduce concepts relevant to the structure of an MDP. We then formalize the nearest neighbor function approximator and show its theoretical guarantee in terms of sample complexity.

\subsection{Metric Space and Intrinsic Dimension}
In practice, the state-action space $\mathcal{X} = \mathcal{S} \times \mathcal{A}$ is usually continuous. We assume that $\mathcal{X}$ is a metric space with a distance function $d_\mathcal{X}$ that satisfies the triangular inequality. This assumption is easily achievable, e.g., the Euclidean distance can be applied to a space of pixel data. We also assume that $\mathcal{X}$ is bounded and has diameter $D := \sup_{x, x'\in \mathcal{X}} d_{\mathcal{X}}(x,x')$.

An intuitive way to measure the complexity of $\mathcal{X}$ is through the ambient dimension $p$, which can be roughly understood as the number of variables \textit{used} to describe a point in $\mathcal{X}$. In real-world MDPs, the states are often represented by real-valued vectors, which form a Euclidean ambient space. Thus, in our context, we simply take $p$ as the Euclidean dimension of the \textit{natural} embedding of $\mathcal{X}$. For instance, a $20 \times 20$ image has $p = 400$ regardless of its content. 

However, most meaningful, high-dimensional data do not uniformly fill in the space where they are represented. Rather, they concentrate on smooth manifolds with low \textit{intrinsic} dimension. The intrinsic dimension $d$ measures the inherent complexity of a metric space. Informally, it is the number of variables \textit{needed} to describe $\mathcal{X}$. Suppose the aforementioned image depicts a car with $5$ physical properties, then $d$ can be $5$ rather than $400$.
Studies on intrinsic dimension estimation \citep[e.g.,][]{Kgl2002IntrinsicDE, Levina2004MaximumLE} employ more formal definitions. Yet the technical details are out of our scope. We only require $d \leq p$ in general.

For the clarity of later proofs, we outline two concepts that can bound the intrinsic dimension of a metric space.

\begin{definition}[Covering and packing]
\label{def:coverandpacking}
An $\epsilon$-cover of a metric space $\mathcal{X}$ is a subset $\hat{\mathcal{X}} \subseteq \mathcal{X}$ such that for each $x \in \mathcal{X}$, there exists $x' \in \hat{\mathcal{X}}$ with $d_\mathcal{X}(x, x') \leq \epsilon$. The $\epsilon$-covering number is $N(\epsilon) = \min\{|\hat{\mathcal{X}}|: \hat{\mathcal{X}}$ is an $\epsilon$-cover of $\mathcal{X}\}$. An $\epsilon$-packing is $\hat{\mathcal{X}} \subseteq \mathcal{X}$ such that $\forall x, x' \in \hat{\mathcal{X}}$, $d_\mathcal{X}(x, x') > \epsilon$. The $\epsilon$-packing number is $M(\epsilon) = \max\{|\hat{\mathcal{X}}|: \hat{\mathcal{X}}$ is an $\epsilon$-packing of $\mathcal{X}\}$.
\end{definition}

\subsection{Nearest Neighbor Function Approximator}
In the RL setting, the learning algorithm collects a set of observations $B := \{x_i\}_{i 
\in [N]}$ and corresponding value labels $\{Q(x_i)\}_{i 
\in [N]}\subseteq \mathbb{R}$. The unknown $Q:\mathcal{X}\rightarrow \mathbb{R}$ measures the quality of a state-action pair. As in supervised machine learning, the task is to find a function approximator $f: \mathcal{X}\rightarrow \mathbb{R}$ that approximates the known labels with small errors and also generalizes to unseen data points. That is, $f(x_i)\approx Q(x_i)$ for all $i\in [N]$ and $f(x)\approx Q(x)$ for $x \in \mathcal{X}\backslash B$.
With the distance metric $d_\mathcal{X}$, we can now define the NN approximator which satisfies the above property.

\begin{definition}[\textbf{Nearest neighbor function approximator}]
\label{def:NNApproximator}
Given a sample buffer $B = \{\big(x_i, Q(x_i)\big)\}_{i 
\in [N]} \subseteq \mathcal{X} \times \mathbb{R}$, 
the nearest neighbor approximator is the function $\hat{Q}: \mathcal{X} \rightarrow \mathbb{R}$ such that $\forall x \in \mathcal{X}$,
\begin{align*}
    \hat{Q}(x) := \min_{i \in [N]}\{Q(x_i) + L \cdot d_\mathcal{X}(x, x_i)\}.
\end{align*}
$L>0$ is a parameter that adjusts the approximation error.
\end{definition}

Note that $\hat{Q}(x)$ matches existing samples exactly and the approximation error for a new $x$ is characterized by an upper bound obtained from the closest known data. This contrasts with other function approximators that lack theoretical understanding, e.g., neural networks. Now, we proceed to show more practical guarantees of the NN function approximator.

\subsection{MDP with Lipschitz Continuity}
To ensure the problem is tractable, we impose the following regularity condition on the optimal Q-function of the MDP. 

\begin{assumption}[MDP with Lipschitz continuity]
\label{assumption:LipMDP}
In the metric space $\mathcal{X}$, let the optimal Q-function be $Q^*_h: \mathcal{X} \rightarrow \mathbb{R}$, then $\exists L_1$, $L_2 > 0$ such that $\forall x$, $x' \in \mathcal{X}$, $\forall h \in [H]$,
\begin{align}
    |Q^*_h(x) - Q^*_h(x')| \leq L_1 \cdot d_\mathcal{X}(x, x'),
    \label{QLip}
\end{align}
\begin{align}
    \max_{a''} d_\mathcal{X}\big[\big(f(x), a''\big), \big(f(x'), a''\big)\big] \leq L_2 \cdot d_\mathcal{X}(x, x').
    \label{distLip}
\end{align}
\end{assumption}
Assumption \ref{assumption:LipMDP} implies that there is a proper notion of distance in $\mathcal{X}$ such that two points close to each other have similar Q-values. It also ensures a stable system. Let $L = L_1(L_2 + 1)$ be the parameter defining $\hat{Q}(x)$, then $\hat{Q}(x)$ is $L$-Lipschitz continuous \cite[Lemma~2]{yang2019learning}. 

\citet{yang2019learning} proposed an upper-confidence algorithm with NN approximation (UCRL-FA) for discrete deterministic MDPs. Basically, the approximate $Q_h$ is updated recursively from $h = H$ to $h = 1$ at the end of each episode. The agent acts according to the greedy policy $\argmax_{a \in \mathcal{A}} Q_h$. For completeness, we present UCRL-FA in Appendix~A. It achieves the following regret bound.

\begin{theorem}[Regret till $\epsilon$-optimality by Yang et al]
\label{theorem:Yang}
Suppose $\mathcal{X}$ admits an $\epsilon$-cover with size $N(\epsilon)$ for any $\epsilon > 0$. After $K$ episodes, UCRL-FA with the nearest neighbor construction obtains a regret bound $Regret(K) \leq HN(\epsilon) + 2\epsilon LKH.$ If $\mathcal{X}$ is compact, then $Regret(K) = O(DLK)^\frac{p}{p+1} \cdot H$.
\end{theorem}

For a metric space with ambient dimension $p$, Theorem~\ref{theorem:Yang} states that UCRL-FA learns the system with $\epsilon^{-p}$ samples, where $\epsilon$ is the learning accuracy.
The algorithm is efficient for low-dimensional observations.  Indeed, in real-world MDPs governed by laws of physics, the intrinsic dimensions are usually small. However, the internal states are not always
available. If there are only pixel data, $p$ can be hundreds if not thousands even for a simple system like a moving car. The natural question is: \textit{can the NN function approximator achieve a regret that depends only on the intrinsic dimension $d$?}
In the next section, we answer the question affirmatively with some mild assumptions.

\subsection{Efficient Value Learning in Metric Spaces}
To characterize the true complexity of the NN approximator, we additionally distinguish the state-action spaces.
Suppose the learner observes a $p$-dimensional state space $\hat{\mathcal{S}}$, e.g., image space, which admits an  embedding to a $d$-dimensional space $\mathcal{S}$, e.g., the parameter space of a physical system,
for some $d\ll p$.
We emphasize that the learner does not have any information about $\mathcal{S}$.
Let $\mathcal{Y} = \hat{\mathcal{S}} \times \mathcal{A}$ and $d_\mathcal{Y}$ be the distance metric satisfying the triangular inequality.
Since the action space remains the same, we simply denote the ambient dimension of $\mathcal{Y}$ as $p$ and the intrinsic dimension as $d$.

The relationship between any true state $s$ and its external representation $\hat{s}$ can be described by a function that maps $\mathcal{S}$ to $\hat{\mathcal{S}}$.  In our proofs, we only assume the existence of the mapping, but do not require knowing its explicit form.

\begin{assumption}[Bi-Lipschitz mapping between metric spaces]
\label{assumption:BiLip}
There exists a map $g : \mathcal{X} \rightarrow \mathcal{Y}$ from the intrinsic state space to the external metric space. We assume that $g$ is Bi-Lipschitz, i.e., $\forall x, x' \in \mathcal{X}$, $\exists C > 0$ such that 
\begin{align}
    C^{-1} d_\mathcal{X}(x, x') \leq d_\mathcal{Y}(g(x), g(x')) \leq C d_\mathcal{X}(x, x').
    \label{BiLip}
\end{align}
$d_\mathcal{X}$ and $d_\mathcal{Y}$ are distance metrics for $\mathcal{X}$ and $\mathcal{Y}$, respectively.
\end{assumption}

Assumption \ref{assumption:BiLip} models most real-world RL systems: if two points are close in the observation space, they are close internally. Now, 
by Assumption \ref{assumption:LipMDP}, we obtain the following lemma on the continuity of the observation space. 
\begin{lemma}[High-dimensional MDP with Lipschitz continuity]
\label{lemma:HighDimMDP}
If $\mathcal{X}$ satisfies Assumption \ref{assumption:LipMDP} and $g : \mathcal{X} \rightarrow \mathcal{Y}$ satisfies Assumption \ref{assumption:BiLip}, then the MDP with state-action space $\mathcal{Y}$ is $\hat{L}$-Lipschitz continuous, where $\hat{L} = (L_2C^2 + 1) \cdot L_1C$.
\end{lemma}

The formal proof can be found in Appendix~B.
Now, we can treat $\mathcal{Y}$ as an independent MDP without knowing its intrinsic properties and directly apply Theorem \ref{theorem:Yang}. 
Lemma~\ref{lemma:LocalRegret} characterizes the regret of the NN approximator in $\mathcal{Y}$.
\begin{lemma}[Regret in $\mathcal{Y}$ w.r.t. the local dimension]
\label{lemma:LocalRegret}
Suppose that $\mathcal{Y}$ admits an $\hat{\epsilon}$-cover $N(\hat{\epsilon})$ for some $\hat{\epsilon} > 0 $, after $K$ episodes, the NN function approximator obtains a regret
\begin{align}
    Regret(K) \leq H \cdot N(\mathcal{Y}, \hat{\epsilon}) + 2\hat{\epsilon}\hat{L}KH
    \label{YRegBound}
\end{align}
where $\hat{L} = (L_2C^2 + 1) \cdot L_1C$.
\end{lemma}

To bound the regret of the NN approximator in the observation space, it remains to bound the covering size of $\mathcal{Y}$. Recall the covering and packing of a metric space in Definition \ref{def:coverandpacking}, we now present the main theorem.
\begin{theorem}[\textbf{Regret till $\hat{\epsilon}$-optimality in $\mathcal{Y}$ w.r.t. the intrinsic dimension}]
\label{theorem:IntrinsicRegret}
Suppose $\mathcal{X}$ admits an $\epsilon$-cover $N(\mathcal{X}, \epsilon)$ and $g: \mathcal{X} \rightarrow \mathcal{Y}$ satisfies Lemma \ref{lemma:HighDimMDP}, then the following holds:
\begin{enumerate}
    \item $\mathcal{Y}$ admits an $\hat{\epsilon}$-cover with $\hat{\epsilon} = 2C\epsilon$, $N(\mathcal{Y}, \hat{\epsilon}) \leq N(\mathcal{X}, \epsilon)$
    \item In $K$ episodes, UCRL-FA with NN function approximator obtains a regret bound $O(D\hat{L'}K)^\frac{d}{d+1} \cdot H$ with $\hat{L'} := C\hat{L}$. 
\end{enumerate}
\end{theorem}

\begin{proof}
We make use of the facts that
\begin{align}
    \forall \epsilon > 0, M(2\epsilon) \leq N(\epsilon) \leq M(\epsilon)
    \label{theorem:DimInequalities}
\end{align}
and for a metric space with diameter $D$ and intrinsic dimension $d$, there exists a $\epsilon$-cover with size 
\begin{align}
    N(\epsilon) = \Theta\left(D/\epsilon\right)^d. \label{theorem:DimBound}
\end{align}

We want to show that the covering number of $\mathcal{Y}$ cannot be greater than the covering number of $\mathcal{X}$ when $\hat{\epsilon}$ is set to $2C\epsilon$. Then, the regret bound in \eqref{YRegBound} can be replaced by an upper bound which depends on the inherent properties of $\mathcal{Y}$.

Note that finding $M(\epsilon)$ for a dataset $B_n = \{x_1, ..., x_n\}$ is equivalent to finding the cardinality of a maximum independent set $MI(G_\epsilon)$ in the graph $G_\epsilon(V,E)$ with vertex set $V = B_n$ and edge set $E = \{(x_i,x_j)|d(x_i,x_j) < \epsilon\}$.

Now, consider the graph $G^\mathcal{X}_{2\epsilon}$ constructed by the above rule. Any two connected points $x, x'$ in the graph satisfy $d_\mathcal{X}(x, x') < 2\epsilon$. Denote the image of a maximum independent set $MI_\mathcal{X}(G^\mathcal{X}_{2\epsilon})$ in $\mathcal{Y}$ as $MI_\mathcal{Y}(G^\mathcal{X}_{2\epsilon})$. $MI_\mathcal{Y}(G^\mathcal{X}_{2\epsilon})$ is still a maximum independent set w.r.t. the graph with vertex set $V = B_n$ and edge set $E_\mathcal{X} = \{(g(x_i),g(x_j))|d_\mathcal{X}(x_i,x_j) < 2\epsilon\}$ in the new metric space. 

To find the packing number of $\mathcal{Y}$, we require the cutoff condition for $E_\mathcal{Y}$ to be $d_\mathcal{Y}[g(x_i),g(x_j)] < 2C\epsilon$. By \eqref{assumption:BiLip}, $d_\mathcal{Y}[g(x),g(x')] < 2C\epsilon$ for all elements in $E_\mathcal{X}$. Therefore, $E_\mathcal{X} \subseteq E_\mathcal{Y}$. In other words, $g(G^\mathcal{X}_{2\epsilon})$ is a subgraph of $G^\mathcal{Y}_{2C\epsilon}$ with the same vertices. Thus, $MI_\mathcal{Y}(G^\mathcal{Y}_{2C\epsilon}) \leq MI_\mathcal{X}(G^\mathcal{X}_{2\epsilon})$. This is because adding edges can only reduce (or remain) the size of the maximum independent sets in a graph. 

Thus, $M(\mathcal{Y}, 2C\epsilon) \leq M(\mathcal{X}, 2\epsilon)$. Using \eqref{theorem:DimInequalities}, we conclude that $N(\mathcal{Y}, 2C\epsilon) \leq M(\mathcal{Y}, 2C\epsilon) \leq M(\mathcal{X}, 2\epsilon) \leq N(\mathcal{X}, \epsilon)$, where the first inequality holds in space $\mathcal{Y}$, the second inequality comes from the proofs above, and the last inequality holds in space $\mathcal{X}$. Preserving only the first and the last terms, we have $N(\mathcal{Y}, \hat{\epsilon}) \leq N(\mathcal{X}, \epsilon)$. 

Consequently, the regret upper bound in \eqref{YRegBound} becomes 
\begin{align*}
    Regret(K) \leq H \cdot N(\mathcal{Y}, \hat{\epsilon}) + 2\hat{\epsilon}\hat{L}KH \\\leq H \cdot N(X, \epsilon) + 4C\hat{L}\epsilon KH.
\end{align*}

In $\mathcal{X}$, equation \eqref{theorem:DimBound} indicates that $N(\mathcal{Y}, \hat{\epsilon}) \leq \Theta(D/\epsilon)^d$, where $d$ and $D$ are the intrinsic properties of the state-action space. As a result, $Regret(K) \leq  H\cdot \Theta(D/\epsilon)^d + 4C\hat{L}\epsilon KH$.

Denote $\hat{L'} := C\hat{L}$ as the smoothing constant for the high-dimensional metric space, when $
    \epsilon = D^\frac{d}{d+1} \cdot (CLK)^{-\frac{1}{d+1}}
$,
the upper bound becomes 
$
O(D\hat{L'}K)^\frac{d}{d+1} \cdot H
$ as desired.
\end{proof}

The main takeaway is that the regret in the complex space is also sub-linear in $K$ and linear in $H$. Moreover, it depends on the intrinsic dimension rather than the ambient dimension. In practice, the NN approximator makes learning from images as efficient as from the actual state descriptors by emphasizing the internal differences between observations.

\section{Nearest Neighbor Actor-Critic}
In this section, we integrate the NN approximator into the actor-critic framework and evaluate its practicality. This Nearest Neighbor Actor-Critic (NNAC) combines a policy network and an NN critic to solve RL problems. Upon receiving a new observation, the NN critic finds a sequence of past $(s, a)$ pairs as a simulated trajectory and sums up the upper-bounded rewards as the value estimate. The 1-step TD error is obtained from consecutive state values. Then, the policy is updated based on the log action probability scaled by the TD error. The pseudocode is given in Algorithm \ref{algo:NNAC}.

\begin{algorithm}[t!]
\footnotesize
\caption{Nearest Neighbor Actor-Critic}
\begin{algorithmic}[0]
\State Initialize experience buffer $B^{1} = \emptyset$, policy parameter $\theta_{\pi^1}$

\For{$k = 1,..., K_{max}$}
    \State Receive initial state $s^k_1$
    \For{$h = 1,..., H$}
        \State Take action $a^k_h$ according to policy $\pi^k(a^k_h | s^k_h, \theta_{\pi^k})$
        \State Receive $r^k_h \leftarrow r(s^k_h, a^k_h)$ and $s^k_{h+1} \leftarrow f(s^k_h, a^k_h)$
        
        \State $\hat{V}(s^k_h) \leftarrow$ \Call{NNFuncApprox} {$s^k_h, h, \pi^k, B^k, H$}
        
        \State $\hat{V}(s^k_{h+1}) \leftarrow$  \Call{NNFuncApprox}{$s^k_{h+1}, h+1,  \pi^k,  B^k, H$}
        \State $\delta^k_h = r(s^k_h,a^k_h)+\gamma\cdot \hat{V}(s^k_{h+1})- \hat{V}(s^k_h)$
        
        \State $B^k \leftarrow B^k \cup \{\big(s^k_h, a^k_h, f(s^k_h, a^k_h), r(s^k_h, a^k_h), \delta^k_h\big)\}$
        
        \State Sample a random mini-batch of $N$ transitions from $B$
        \State Update policy with TD error policy gradient by Eq. \eqref{eq:PolicyGrad}
    \EndFor
    \State \textbf{end for}
    \State $B^{k + 1}\leftarrow B^k$
\EndFor
\State \textbf{end for}
\hrulefill
\Procedure{NNFuncApprox}{$s$, $h$, $\pi(\cdot|\theta_\pi)$, $B$, $H$}
    \If{$h == H$}
        \State \Return 0
    \EndIf
    \State $\mathcal{V} = \emptyset$
    \State $a \leftarrow \pi(s|\theta_\pi)$
    \For{$i = 1, ..., M$}
        
        \State $(s_i, a_i) \leftarrow i$-th neighbor of $(s, a)$ in $B$ under metric $d$
        \State $r_i \leftarrow r(s_i, a_i)$, $s'_i \leftarrow f(s_i, a_i)$ \Comment{stored in $B$}
        \State $\hat{V}'_i \leftarrow$ \Call{NNFuncApprox}{$s'_i, h+1,\pi(\cdot|\theta_\pi), B, H$}
        \State $\hat{V}_i \leftarrow r_i + \gamma \cdot \hat{V}'_i + L \cdot d[(s, a), (s_i, a_i)]$
        \State $\mathcal{V} \leftarrow \mathcal{V} \cup \{\hat{V}_i\}$
    \EndFor
    \State \textbf{end for}
    \State \Return $\min\mathcal{V}$
\EndProcedure
\end{algorithmic}
\label{algo:NNAC}
\end{algorithm}

\subsection{Neural-Network-Based Actor}
Unlike the aforementioned tabular methods which employ the greedy policy $\pi (s) = \argmax_a \hat{Q}(s,a)$, we use a separate network parameterized by $\theta_\pi$ to guide the actor’s movement. Let the actor loss be $J(\theta_\pi)$. The network is updated through the standard TD error policy gradient \cite{Sutton1988LearningTP} with mini-batch size $N$:
\begin{align}
    \quad\;\;\nabla_{\theta_{\pi}}J = N^{-1} \sum
        \delta_i \nabla_{\theta_\pi}\log\pi(a_i | s_i, \theta_{\pi}).
    \label{eq:PolicyGrad}
\end{align}
At step $h$, the action distribution at $s_h$ is pushed towards $a_h$ if the TD error $\delta_h > 0$. We use temporal difference learning instead of directly maximizing the values to reduce variance.

\subsection{Nearest-Neighbor-Based Critic}
The NN critic in Algorithm \ref{algo:NNAC} maintains an experience buffer $B$ to construct the nearest neighbor tree. As in \cite{Schulman2016HighDimensionalCC}, we use a parameter $\gamma \in (0, 1]$ to down-weight future rewards due to delayed effects. This parameter corresponds to the discount factor in infinite-horizon discounted problems, but we take it as a variance reduction technique.

Following previous notation, let $x_h$ denote a state-action pair at step $h$. Let $M$ be the number of neighbors considered in value approximation. The exact estimation procedure exploits Monte Carlo tree search. Given $x_h$, find $M$ samples in history with the minimum distances $d_\mathcal{X}(x_h, x_i)$, $i \in [M]$. Consider their next states and the actions that would be taken under $\pi_h$. Expand these new state-action pairs until the tree depth reaches $H - h$. The total reward is back-propagated from the terminal states and the minimum value is the estimate for $V^\pi(x_h)$. When $M = 1$, the recursive formula is:
\begin{align} 
    \hat{V}^\pi(x) = r(x) + \gamma \cdot \hat{V}^\pi(x') + L \cdot d_\mathcal{X}(x, x').
    \label{eq:Recursion}
\end{align}
$x' = \argmin_{x' \in B}d_\mathcal{X}(x, x')$ is the nearest neighbor and $\hat{V}^\pi(x') = \hat{Q}^\pi\big(x', \pi(x')\big)$. In long-horizon problems, we can replace the varying $H - h$ with a fixed planning horizon $H'$.

Equation \eqref{eq:Recursion} prioritizes exploring $(s, a)$’s that are farther away from the seen ones. By Assumption \ref{assumption:LipMDP}, the Lipschitz bonus $L \cdot d_\mathcal{X}(x, x')$ ensures that the real value of $V^\pi(x)$ is upper-bounded by $\hat{V}^\pi(x)$. As new observations accumulate, $d_\mathcal{X}(x, x')$ becomes smaller and the upper bound is improved. Since exploration is based on the value upper bound, NNAC encourages exploring new actions. In training, $d_\mathcal{X}$ is problem-specific and $L$ can be tuned as a hyperparameter.

The distance-based bonus is less sensitive to the dimension of $\mathcal{X}$. Deep RL has poor sample efficiency for pixel data as it makes implicit use of feature encoding to find a low-dimensional embedding. In contrast, the NN approximator does not need any parametric model to capture the intrinsic states. With a proper metric defined in the high-dimensional space, the neighbors preserve their similarities and the distances can still be good indicators of their relative values.

\subsection{Evaluation of NNAC}
We test NNAC on the cart-pole balancing problem.
Due to the discrete nature of the action space, we compare with dueling double deep Q-networks (DDDQN) \cite{Wang2016DuelingNA}, trust region policy optimization (TRPO) \cite{schulman2015TRPO}, proximal policy optimization (PPO) \cite{Schulman2017ProximalPO}, soft actor-critic (SAC) \cite{Haarnoja2018SoftAO}, and neural episodic control (NEC) \cite{Pritzel2017NeuralEC}. Our goal is to show that the NN critic enables efficient learning with high-dimensional data, which is generally not easy to achieve. Therefore, we select a task with relatively simple dynamics and do not compare with all state-of-the-art deep RL methods that typically use more samples.

\subsubsection{Cart-pole environment:} We use the OpenAI Gym implementation \cite{Brockman2016OpenAIG}.  The state of the cart is described by a 4-tuple $(\theta, \Dot{\theta}, x, v)$, where $\theta$ and $\Dot{\theta}$ are the angle and angular velocity of the pole, $x$ is the horizontal position of the cart, and $v$ is the velocity. The horizon and the maximum achievable reward in one episode are both $500$.

\subsubsection{State space dimension:} We prepare three types of inputs. 

\begin{itemize}
    \item dim$(\mathcal{S}) = 4$. The 4-tuple descriptors are used directly. 

    \item dim$(\mathcal{\hat{S}}) = 10$ and $100$. We use random projection matrices to map the 4-tuples into high-dimensional spaces. The matrix columns are orthonormal to preserve the distances and the neighbor relations in the new metric spaces.
    
    \item dim$(\mathcal{\hat{S}}) = 4 \times 20 \times 20$. We crop the cart-pole from the $160 \times 240$ gray scale images and down-sample it to $20 \times 20$ pixels. Four consecutive frames are stacked together to derive the velocity and acceleration of the moving object. 
\end{itemize}
In all cases, $L_2$ distance is used for the nearest neighbor search. As the $L_2$ distance may not be a good measurement for image similarity, we also learn a distance oracle with the Siamese network for comparison (details in Appendix~C.2).

\begin{figure}[t!]
    \centering
    \begin{subfigure}[b]{0.46\textwidth}
    \includegraphics[width=\textwidth]{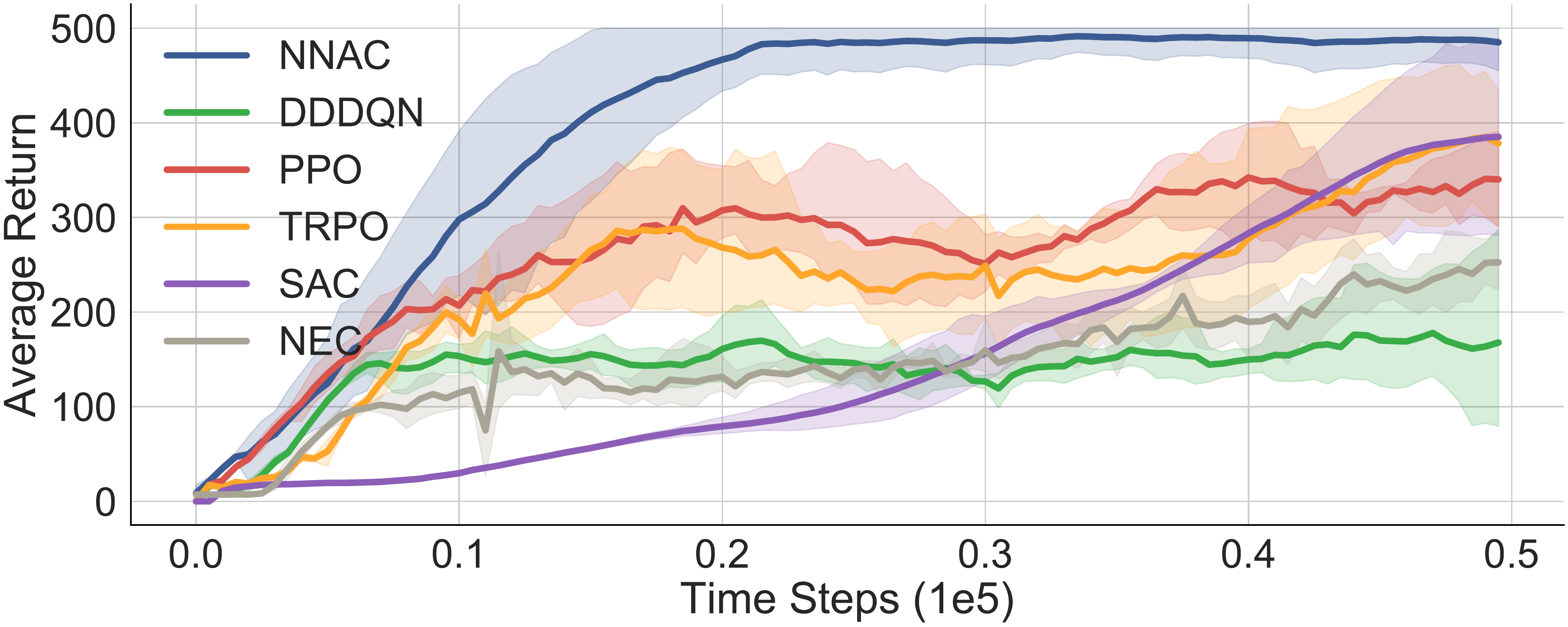}
    \caption{NNAC vs baselines with internal state descriptors}
    \label{fig:NNACcompare}
    \end{subfigure}
    \centering
    \begin{subfigure}[b]{0.46\textwidth}
    \includegraphics[width=\textwidth]{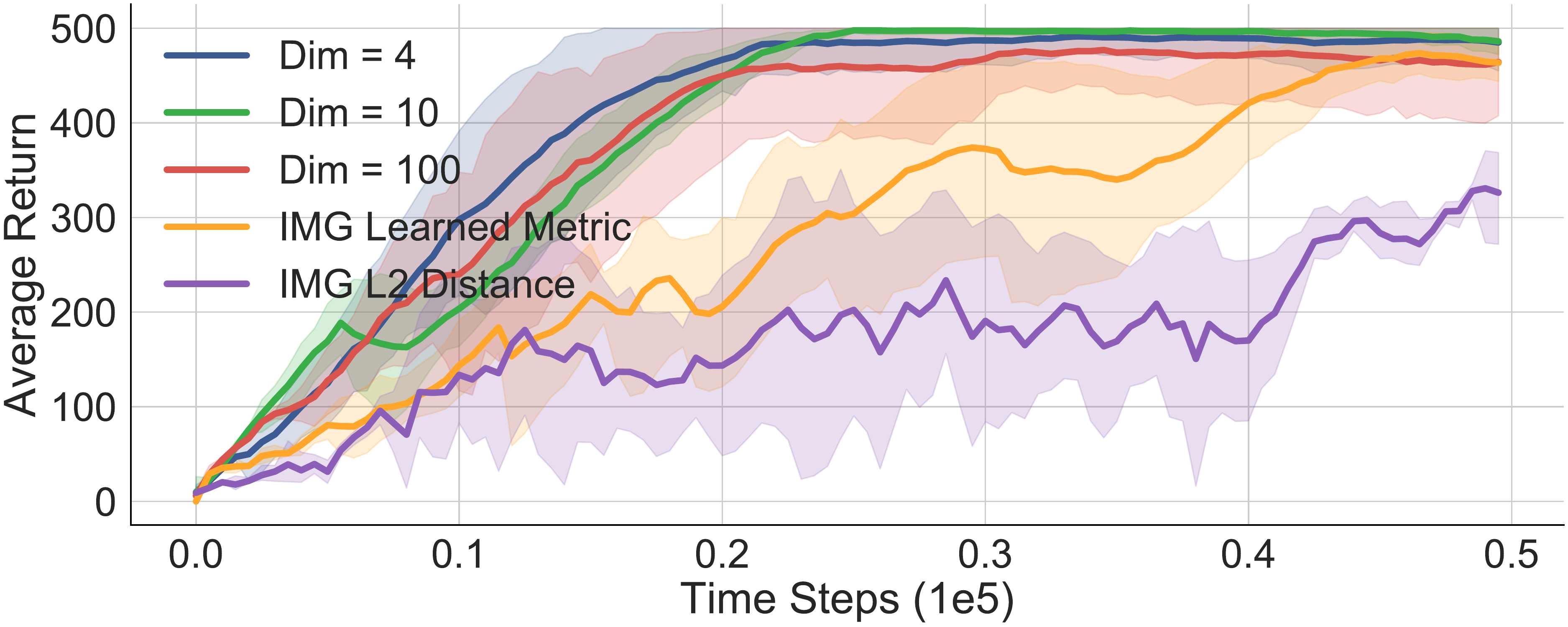}
    \caption{NNAC with different input dimensions}
    \label{fig:NNACdim}
    \end{subfigure}
    \caption{Learning curves for CartPole-v1. The shaded areas denote one standard deviation of evaluations over 5 trails. Curves are smoothed by taking a $500$-step moving average.}
    \label{fig:cartpole}
\end{figure}

\subsubsection{Network structure and hyperparameters:} We use the Stable Baselines implementation of the deep RL agents \cite{stable-baselines}. For low-dimensional NNAC, we use a one-layer policy network with 32 hidden units and a ReLU activation. When learning from pixels, we add a convolutional layer with 16 units before the policy network. The discount factor $\gamma$ is $0.99$. The Lipschitz $L$ is determined by a grid search and set to $7$. All agents are trained with $5$ random seeds. Evaluation is done every $1000$ steps without exploration. 

\subsubsection{Results and discussion:}
Figure \ref{fig:NNACcompare} shows the learning curves for all agents when the internal states are given. NNAC learns better policies with fewer samples than the other baselines. Also, compared with network critics that need extensive hyperparameter tuning, the NN approximator only requires experimenting with $L$. 

Figure \ref{fig:NNACdim} illustrates the performance of NNAC with different input dimensions. It converges in similar steps for the intrinsic and projected states. This agrees with our proposition since the distances are preserved by the matrix transformation. For the pixel data coupled with a learned metric, NNAC achieves the same performance with slightly more samples, which might be related to learning the convolutional filters. When $L_2$ distance is used to measure image similarities, the algorithm is less stable and does not solve the environment within limited time steps. However, the final average return is already comparable to the deep RL agents with internal state input. Indeed, deep RL typically uses a linear output layer after several nonlinear layers. It can be interpreted as a feature encoding process followed by linear value approximation. Rather than learning the encoding, NNAC treats the distance metric as known information, thereby reducing the amount of unnecessary work. 

Empirically, we find the $L_2$ distance to be a good choice for low-dimensional, physical state spaces, but other sophisticated metrics might be needed to capture the differences between images. For generic high-dimensional tasks, our algorithm is efficient as long as a distance oracle is provided.

\begin{algorithm}[t!]
\caption{Soft Nearest Neighbor Update}
\footnotesize
\begin{algorithmic}[0]
\State // Assume a small constant $\epsilon > 0$
\State Initialize policy network $\theta_\pi$, value network $\theta_Q$, $B = \emptyset$, $\alpha \leftarrow \alpha_0$

\For{each episode}
    \For{each environment step}
        \State Take action $a$ according to $\pi(s)$ and exploration strategy
        \State Receive reward $r$ and next state $s'$
        \If {$\alpha > \epsilon$}
            \State $\hat{V}(s) \leftarrow$ \Call{NNFuncApprox}{$s, h, \pi, B, H$}
            \State $\hat{V}(s') \leftarrow$ \Call{NNFuncApprox}{$s', h+1,  \pi,  B, H$} 
            \State $\delta = r+\gamma\cdot \hat{V}(s')- \hat{V}(s)$
        \EndIf
        \State \textbf{end if}
        \State $B \leftarrow B \cup \{(s, a, s', r, \delta)\}$
    \EndFor
    \State \textbf{end for}
    \For{each gradient step}
        \State Update the actor by Equation \eqref{equ:modifedActorLoss}
        \State Update the critic by Equation \eqref{equ:modifedCriticLoss}
    \EndFor
    \State \textbf{end for}
    \State $\alpha \leftarrow (1 - \beta)\cdot \alpha$
\EndFor
\State \textbf{end for}
\end{algorithmic}
\label{algo:framework}
\end{algorithm}

The major concern of NNAC is the computational cost of finding the neighbors. We use Kd-tree \cite{Friedman1977AnAF} in our implementation. For more complicated MDPs, the training data size can be too large to build a Kd-tree efficiently. Thus,
in the next section, we introduce a new method which preserves the original neural networks of deep RL agents to reduce the computational burden.

\begin{figure*}[!t]
    \centering
    \begin{subfigure}[b]{0.24\textwidth}
        \includegraphics[width=\textwidth]{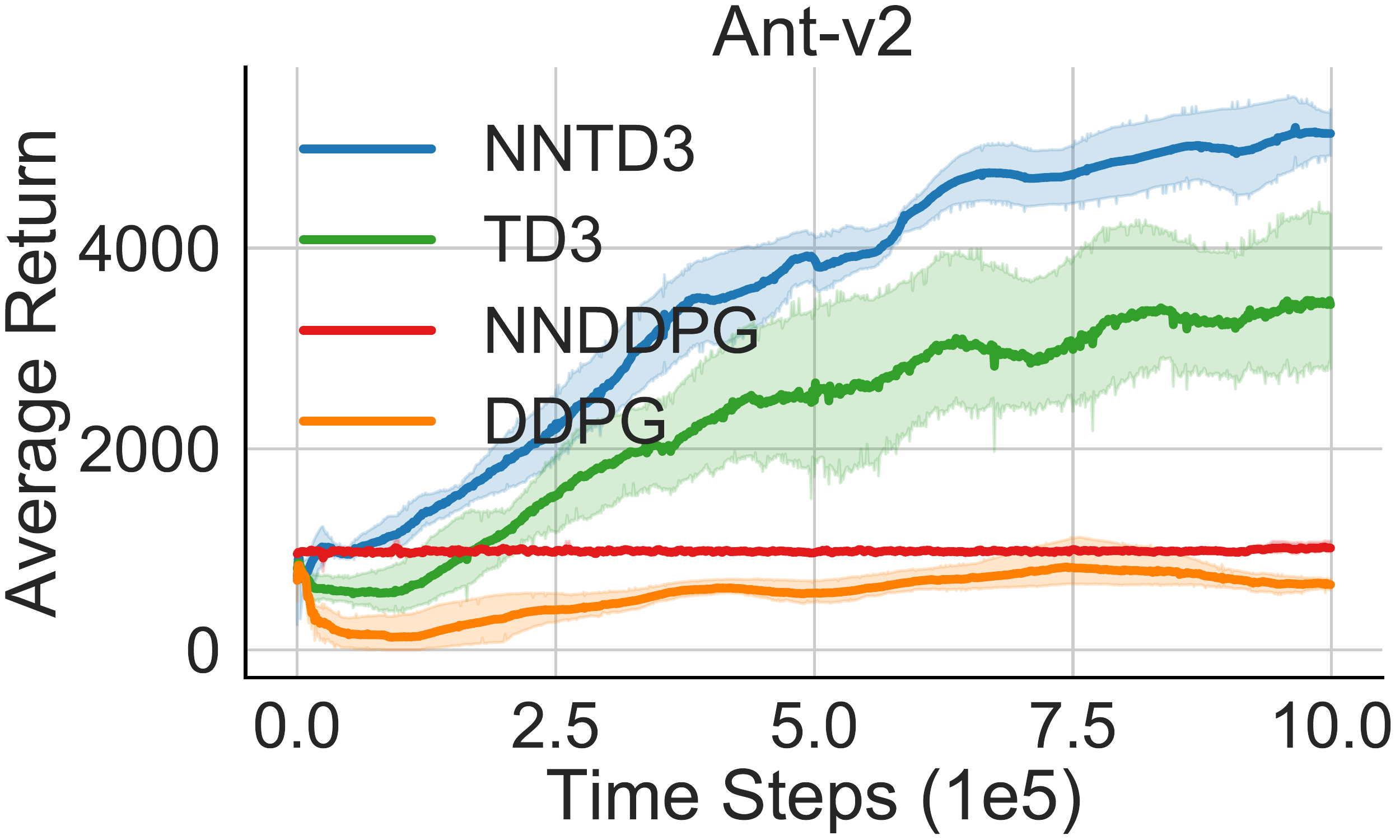}
    \end{subfigure}~
    \begin{subfigure}[b]{0.24\textwidth}
        \includegraphics[width=\textwidth]{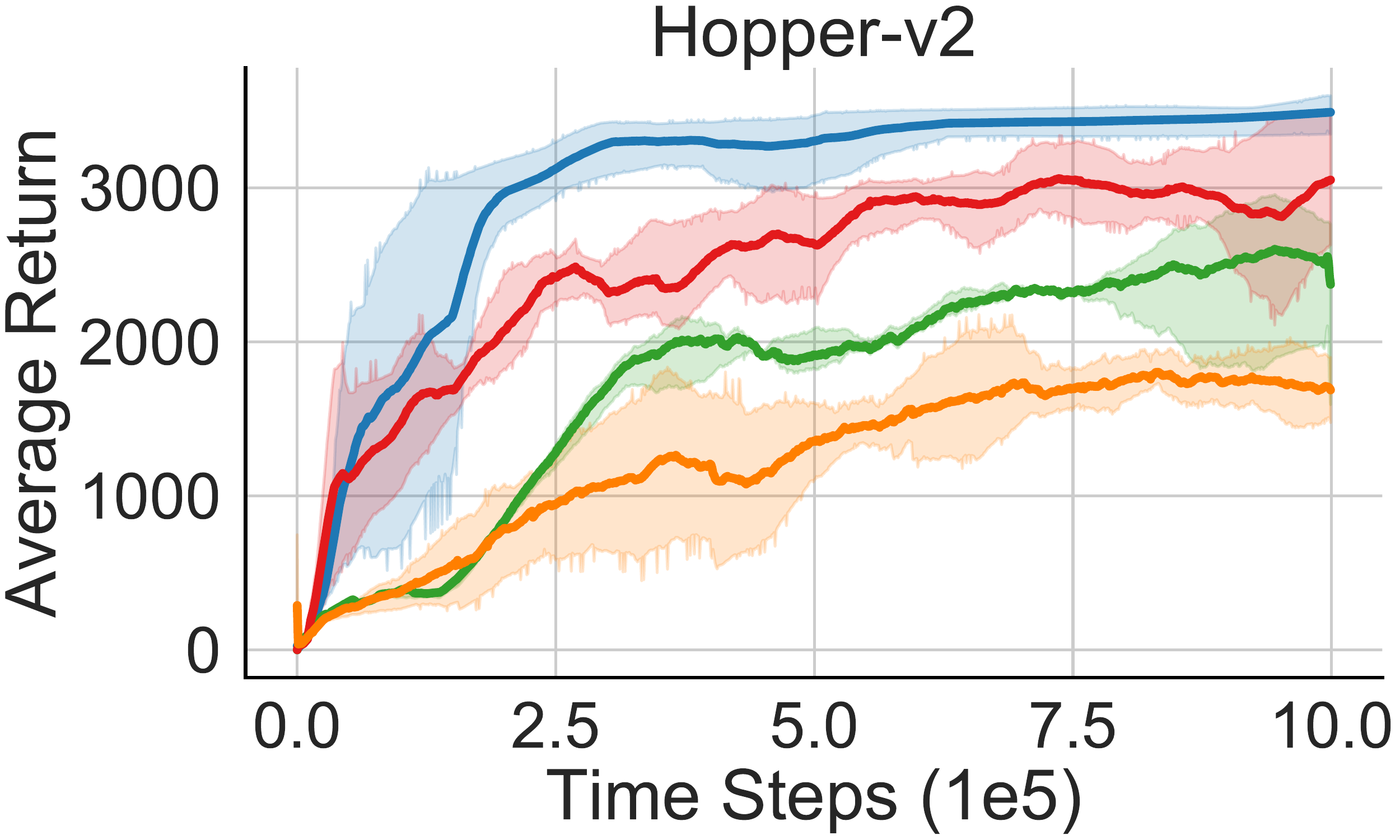}
    \end{subfigure}~
    \begin{subfigure}[b]{0.24\textwidth}
        \includegraphics[width=\textwidth]{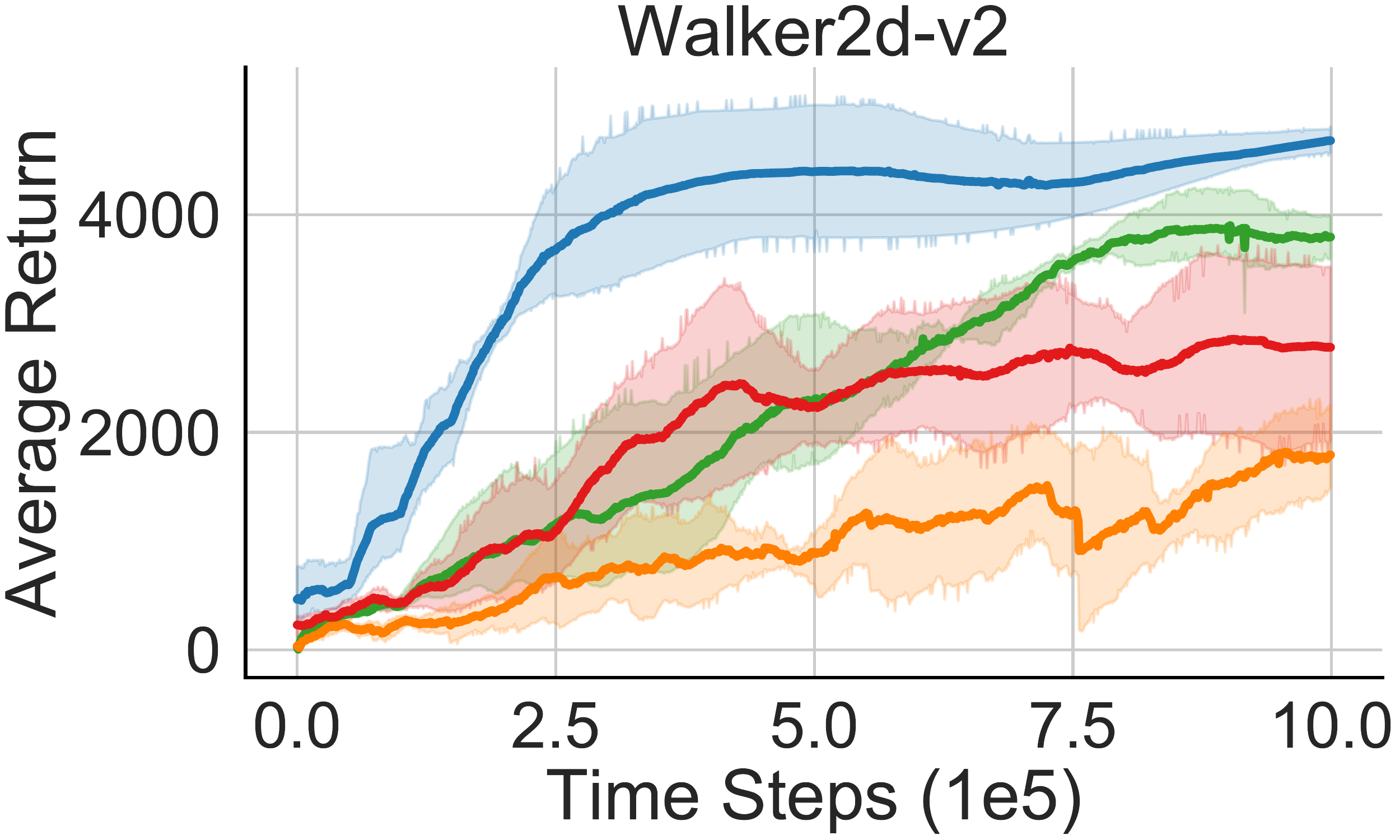}
    \end{subfigure}~
    \begin{subfigure}[b]{0.24\textwidth}
        \includegraphics[width=\textwidth]{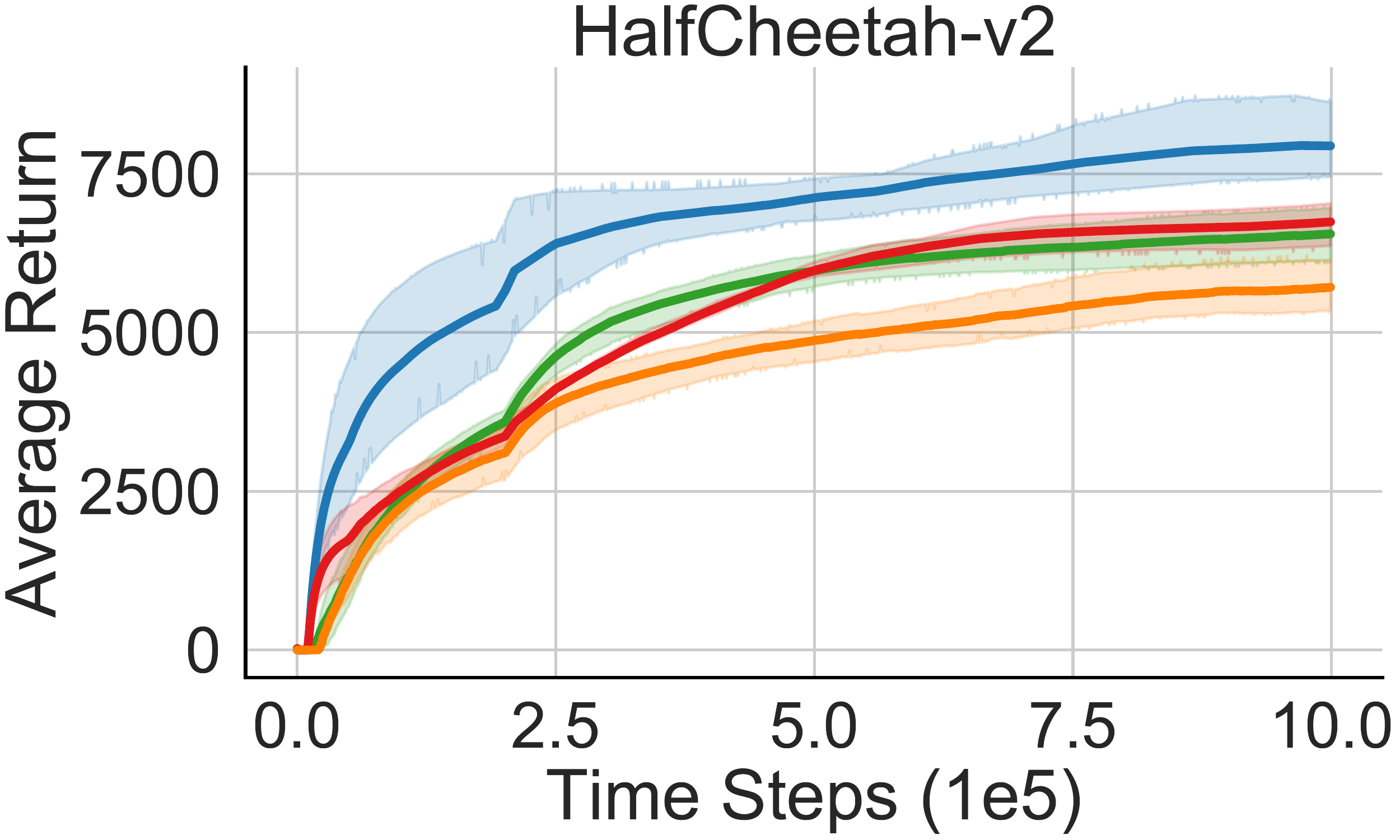}
    \end{subfigure}
    \caption{Learning curves for OpenAI Gym MuJoCo continuous control tasks. The shaded areas denote one standard deviation of evaluations over 5 trails. For visual clarity, the curves are smoothed by taking a moving average of $10^4$ environment steps.
    }
    \label{fig:MuJoCo}
\end{figure*}

\section{Nearest Neighbor Update Module}
In this section, we show that the NN approximator can boost the efficiency of existing deep RL algorithms. As illustrated in previous sections, our method explores the action space efficiently both in theory and in practice. However, as samples accumulate, a neural network with sufficient training data can outperform other function approximators due to its generalization ability.
Hence, we propose a nearest neighbor plug-and-play module (Algorithm \ref{algo:framework}) that 
acts as a ``starter'' to accelerate deep RL agents and can be removed later. 

We focus on actor-critic methods. Without changing the algorithm's structure, a plug-in NN critic supplies value estimates to the rest of the framework. While the actor benefits from TD error policy gradient, the value network in the original algorithm is penalized for large deviations from the NN estimates. 
We adopt an adaptive weighting scheme and decrease the weight for the NN module to avoid computational bottleneck. Let $\alpha$ be the weight of the NN approximator. There are three major components in the update module.

\subsection{Modification 1: NN Value Estimation}
An NN critic is incorporated to guide the training of both the actor and critic networks. In addition to the value network, \textsc{NNFuncApprox} in Algorithm \ref{algo:NNAC} is used to estimate the current state values. Assume that the original critic loss is $L(\theta_Q)$. We penalize large differences between the TD errors estimated by the value network and the NN critic. Using the mean squared error, the critic objective is reformulated as:
\begin{align}
    L'(\theta_Q) = (1 - \alpha) \cdot L(\theta_Q) + \alpha \cdot \left\| \delta_{\theta_Q} - \delta_{NN} \right\|^2,
    \label{equ:modifedCriticLoss}
\end{align}
where $\delta_{\theta_Q} = r + \gamma V'_{\theta_Q} - V_{\theta_Q}$ is obtained from the value net, and $\delta_{NN}$ is the TD error supplied by the NN approximator.

\subsection{Modification 2: TD-Regularized Policy Learning}
Similar to NNAC, we also include a TD error policy gradient term in the actor loss. Let the original actor loss be $J(\theta_\pi)$. The modified gradient $\nabla_{\theta_\pi}J'(\theta_\pi)$ is:
\begin{align}
     (1 - \alpha) \cdot \nabla_{\theta_\pi}J(\theta_\pi) + \alpha \cdot \delta_{NN} \nabla_{\theta_\pi} \log \pi(a|s; \theta_\pi).
    \label{equ:modifedActorLoss}
\end{align}
The auxiliary TD term increases the weights for rewarding actions and decreases the weights for less preferable actions.
\subsection{Modification 3: Continual TD Supervision}
When the MDP has a large intrinsic dimension, it is impractical to compute the nearest neighbors at each recursion step for every mini-batch samples. Therefore, we use an exponentially decaying weight parameter for the NN critic: $\alpha = \alpha_0 \cdot (1 - \beta)^k$, where $\alpha_0 \in (0, 1]$ is the initial weight, $\beta \in (0, 1)$ is the decrease rate, $k$ is the episode number. MC simulation terminates when $\alpha$ is close to 0. However, past TD estimates can still supervise the learning and reduce the chance of network forgetting. This is achieved by storing $\delta_{NN}$ along with the stepwise observation in the experience buffer. For mini-batches sampled at each gradient step, \eqref{equ:modifedCriticLoss} and \eqref{equ:modifedActorLoss} are used to update the networks if $\delta_{NN}$ is available. Otherwise, the original gradients are used.

\subsection{Evaluation of Soft Nearest Neighbor Update}
\subsubsection{Setup:} We implement the NN critic on DDPG \cite{Lillicrap2016ContinuousCW} and TD3 \cite{Fujimoto2018AddressingFA} and test with four MuJoCo locomotion tasks \cite{Todorov2012MuJoCoAP}. For fairness, we use the same hyperparameters for each method before and after adding the NN module. The network structure is selected from the benchmark work \cite{Duan2016BenchmarkingDR} and identical for all agents.
$L_2$ distance is used to find the neighbors. 
More experiment details are given in Appendix~E. 
\begin{table}[t!]
    \centering
    \small{
    \begin{tabular}{ccccc}
    \toprule
    Environment & NNTD3 & TD3 & NNDDPG & DDPG \\
    \midrule
    Ant & \textbf{4727.44} & 3425.07 & 1115.00 & 1025.82 \\
    Hopper & \textbf{3704.40} & 2708.25 & 3202.64 & 2029.65\\
    Walker2d & \textbf{5170.20} & 4260.23 & 2917.25 & 2633.86\\
    HalfCheetah & \textbf{8341.74} & 7045.53 & 7001.52 & 6385.43 \\
    \bottomrule
    \end{tabular}}
    \caption{Average return over 5 trials. Top values are bolded.}
    \label{tab:MuJoCo}
\end{table}

\subsubsection{Results and discussion:}
Figure \ref{fig:MuJoCo} shows the experiment results. The auxiliary NN-critic improves the sample efficiency of DDPG and TD3 in most settings, though we do not see a major performance gain for DDPG in Ant-v2. If DDPG cannot solve an environment, the soft update module is of less help given that the original value networks still play a crucial role in learning. We summarize two principal benefits of the NN update framework.
\begin{enumerate}
    \item NN-guided training encourages exploration and helps the agents to overcome local optima. Table \ref{tab:MuJoCo} shows that NN algorithms obtain larger maximum returns. The Lipschitz bonus highlights exploring unvisited states. This directional exploration is more efficient than random noise. 
    \item The upper-bounded value estimation stabilizes training by preventing overestimation of $Q(s, a)$. This technique has a similar effect to the twin Q-networks in TD3.
\end{enumerate}

\section{Conclusion}
In this paper, we provide a nearest neighbor function approximator for efficient value learning and justify its sample complexity for high-dimensional input in deterministic systems. The NN value estimator can be incorporated in model-free deep RL to encourage exploration and stabilize training. Our work suggests that there is great potential to improve deep RL with non-parametric methods. Future works can explore the benefits of nearest neighbor search in active learning or extend the theories to stochastic environments.

\clearpage
\section{Acknowledgements}
We thank all anonymous reviewers for their insightful comments. LY acknowledges the support from the Simons Institute at Berkeley (Theory of Reinforcement Learning).

\bibliography{reference.bib}

\end{document}


\maketitle
\section{UCRL-FA} 
\label{appendix:UCRLFA}
\begin{algorithm}[h]
\caption{\small UCRL-FA \cite{yang2019learning}}
\footnotesize
\begin{algorithmic}[1]
\State \textbf{Input:} A deterministic metric MDP
\State \textbf{Initialize:} $B^{(0)} \leftarrow \emptyset$, $Q_h^{(0)}(s,a) \leftarrow H$, $\hat{r}^{(0)}(s,a) \leftarrow 1$ for all 
\State $\qquad(s, a) \in \mathcal{S} \times \mathcal{A}$, $h \in [H]$

\For{episode $k = 1, 2, ..., K$}
    \For{step $h = 1, 2, ..., H$}
        \State Current state $s_h^{(k)}$
        \State Play action $a_h^{(k)} = \argmax_{a\in\mathcal{A}}Q_h^{(k)}(s_h^{(k)},a)$
        \State Record $s_{h+1}^{(k)} \leftarrow f(s_h^{(k)}, a_h^{(k)})$, $r_h^{(k)} \leftarrow r(s_h^{(k)}, a_h^{(k)})$
    \EndFor
    
    \State $B^{(k+1)} \leftarrow B^{(k)} \cup$ 
    \State $\qquad\big\{ \big(s_i^{(k)}, a_i^{(k)}, f(s_i^{(k)}, a_i^{(k)}), r(s_i^{(k)}, a_i^{(k)})\big), i\in [H]\big\}$
    
    \State $\hat{r}^{(k+1)} \leftarrow \textsc{FuncApprox}\big(\{(s,a),r(s,a)\}_{(s,a) \in B ^{( k+1)}}\big)$
    \State $Q_H^{(k+1)} \leftarrow \hat{r}^{(k+1)}$
    \State Update $Q_h^{(k+1)}$ recursively:
    \State $\qquad Q_h^{(k+1)} \leftarrow \textsc{FuncApprox}\big(\big\{(s,a), r(s,a) +$ 
    \State $\qquad\qquad \sup_{a'\in\mathcal{A}} Q_{h+1}^{(k+1)}\big(f(s,a),a'\big) \big\}\big)$
\EndFor
\end{algorithmic}
\label{algo:UCRLFA}
\end{algorithm}
We present the pseudocode for upper-confidence reinforcement learning with general function approximator (UCRL-FA) proposed by \citet{yang2019learning}. In their problem setting, the MDPs have discrete state and action spaces. The rewards are assumed to be either 0 or 1. When the function approximator in Algorithm \ref{algo:UCRLFA} takes the form of the nearest neighbor construction, Line~12 to Line~16 become:
{\footnotesize
\begin{align*}
   & \hat{r}^{(k + 1)}(s,a) = \min_{(s',a') \in B^{(k+1)}} \big(r(s',a')+L_1\cdot dist[(s,a),(s',a')] \big),\\
   & Q_H^{(k+1)} \leftarrow \min\big[\hat{r}^{(k + 1)}(s,a), 1 \big],\\
    &Q_h^{(k+1)} \leftarrow \min_{(s',a') \in B^{(k+1)}} \big[r(s',a')+ \sup_{a''\in\mathcal{A}} Q_{h+1}^{(k+1)}\big(f(s',a'),a''\big)\\
    &\qquad\qquad+L_1\cdot dist[(s,a),(s',a')]\big].
\end{align*}
}%

\section{Proof of Lemma~6}
\label{appendix:lemmaproof}
\begin{proof}
Though the state representation in $\mathcal{Y}$ is different from $\mathcal{X}$, the MDPs have the same transition model, reward function, and thus the Q-function. Let $\hat{s} = g(s) \in \mathcal{Y}$. Plug the Bi-Lipschitz condition~(5) in Equations~(3) and (4), we have $\forall s \in \mathcal{S}, a \in \mathcal{A}$:
\begin{align*}
\begin{split}
    |Q^*_h(\hat{s},a)-Q^*_h(\hat{s}',a')| &= |Q^*_h(s,a)-Q^*_h(s',a')|\\ 
    &\leq L_1 d_\mathcal{X}[(s,a), (s',a')] \text{ by Eq. (3)}\\
    &\leq L_1C \cdot d_\mathcal{Y}[(\hat{s},a), (\hat{s}',a')] \text{ by Eq. (5)}
\end{split}
\end{align*}
and
\begin{multline*}
    \;\; \max d_\mathcal{Y}[(f(\hat{s},a),a''),(f(\hat{s}',a'),a'')] \\
    \begin{aligned}
    &\leq C \cdot \max  d_\mathcal{X}[(f(s,a),a''),(f(s',a'),a'')] \text{ by Eq.  (5)}\\
    &\leq L_2C \cdot d_\mathcal{X}((s,a),(s',a')) \text{ by Eq. (4)}\\
    &\leq L_2C^2 \cdot d_\mathcal{Y}[(\hat{s},a), (\hat{s}',a')] \text{ by Eq. (5)}\\
    \end{aligned}
\end{multline*}
Lemma~6 follows.
\end{proof}

\section{Details of the Cart-Pole Experiment}
\label{appendix:cp}
\subsection{Environment Specification} \label{appendix:cpenv}
\begin{figure}[t!]
    \centering
    \includegraphics[width=0.18\textwidth]{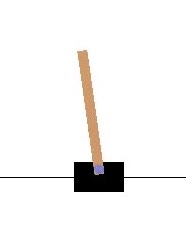}
    \caption{\small{Image of CartPole-v1.}}
    \label{fig:cartpoleimg}
\end{figure}

An example image of the OpenAI Gym CartPole-v1 environment \cite{Brockman2016OpenAIG} is presented in Figure \ref{fig:cartpoleimg}. The state of the cart-pole is a 4-tuple $(\theta, \Dot{\theta}, x, v)$, where $\theta$ is the 
vertical angle of the pole, $\Dot{\theta}$ is the angular velocity, $x$ is the horizontal position of the cart, and $v$ is its velocity. The transition model can be described by the system of equations \cite{Florian2005CorrectEF}:
\begin{equation*}
 \begin{cases}
\Ddot{\theta} = \frac{g\sin{\theta} - \cos{\theta} \frac{F + m_pl\Dot{\theta}^2\sin{\theta}}{m_p+m_l}}{l(\frac{4}{3}-\frac{m_p\cos{\theta}^2}{m_p +m_l})}\\
\Ddot{x} = \frac{F + m_pl(\Dot{\theta}^2\sin{\theta}-\Ddot{\theta}\cos{\theta})}{m_p+m_l}\\
x = x + t \cdot \Dot{x}\\
\Dot{x} = \Dot{x} + t \cdot \Ddot{x}\\
\theta = \theta + t \cdot \Dot{\theta}\\
\Dot{\theta} = \Dot{\theta} + t \cdot \Ddot{\theta}
\end{cases}  
\end{equation*}
An episode ends when either the cart hits the track boundaries ($x \notin [-4.8, 4.8]$) or the pole has fallen over ($\theta \notin [-24^\circ, 24^\circ]$). The agent receives a reward $0$ upon termination and $+1$ otherwise. 

\subsection{Distance Metric Learning} \label{appendix:cpmetricleaerning}
We use the Siamese network to learn the internal distance between two input image stacks. Let the images be $x$ and $x'$ with internal states $s$ and $s'$, respectively. To obtain accurate predictions of the distances, we use larger pixel-size images with dimension $4 \times 160 \times 240$. Therefore in the experiments, we down-sample the rendered images twice with different scales: $160 \times 240$ for distance calculation and $20 \times 20$ for policy learning.

Both $x$ and $x'$ are first passed to a convolutional neural network individually for feature encoding. The network outputs are concatenated together and fed into a fully connected network for distance prediction. Denote the metric network as $\theta_d$. The objective is $L(\theta_d) = \| \theta_d(x, x') -  \| s - s' \|^2 \|^2$. We randomly sample $100000$ $(x, s)$ pairs from the environment and train the network for 5 epochs with learning rate $1 \times 10 ^{-3}$ and batch size 16. The network structure is:
\begin{table}[ht]
\centering
\small
\begin{tabular}{|c|c|}
\hline
Conv 1 & 128 {[}3$\times$3$\times$1{]} filters, leaky ReLU  \\ \hline
Pool   & 2$\times$2 Max with stride 2         \\ \hline
Conv 2 &  64 {[}3$\times$3$\times$1{]} filters, leaky ReLU \\ \hline
Pool   & 2$\times$2 Max with stride 2          \\ \hline
Conv 3 & 16 {[}3$\times$3$\times$4{]} filters, leaky ReLU  \\ \hline
\multicolumn{2}{|c|}{Flatten}\\ \hline
FC     & 64, leaky ReLU    \\ \hline
\multicolumn{2}{|c|}{Concatenate}\\ \hline
FC     & 8, leaky ReLU    \\ \hline
\end{tabular}
\end{table}

\subsection{Hyperparameters for NNAC}
\label{appendix:cphyperparam}
\subsubsection{NN-related parameters}
\begin{itemize}
    \item Number of nearest neighbor $M$: 1
    \item Planning horizon $H'$: 12
    \item Lipschitz parameter $L$: 7, determined by a grid search over $\{0.1, 0.5, 1, 2, ..., 10\}$
    \item Action space dimension: 1
    \item Weight $w$ for $(s, a)$ pairs when calculating distance, i.e., $d(x,x')=\sqrt{\sum w_i(x_i - x'_i)^2}$:
          \begin{itemize}
              \item Dimension 4: $(0.25, 0.25, 0.25, 0.25, 1)$
              \item Dimension 10: $(0.1, ..., 0.1, 1)$
              \item Dimension 100: $(0.01, ..., 0.01, 1)$
              \item Image: $(1, ..., 1, 1)$
          \end{itemize}
          The last dimension is the action weight.
\end{itemize}

\subsubsection{Environment and network-related parameters}
\begin{itemize}
    \item $\gamma$: 0.99
    \item Policy network structure:
          \begin{itemize}
              \item Dimension 4, 10, 100: (32, ReLU, tanh, softmax)
              \item Image: (conv 16, 32, ReLU, tanh, softmax)
          \end{itemize}
    
    \item Network weight initialization: $[-3 \times 10^{-3}, 3 \times 10^{-3}]$
    \item Optimizer: Adam
    \item Mini-batch size: 32
    \item Learning rate: $5\times 10^{-4}$
    
\end{itemize}

\subsection{Evaluation with Pixel Data}

\begin{figure}[h]
    \centering
    \includegraphics[width=0.45\textwidth]{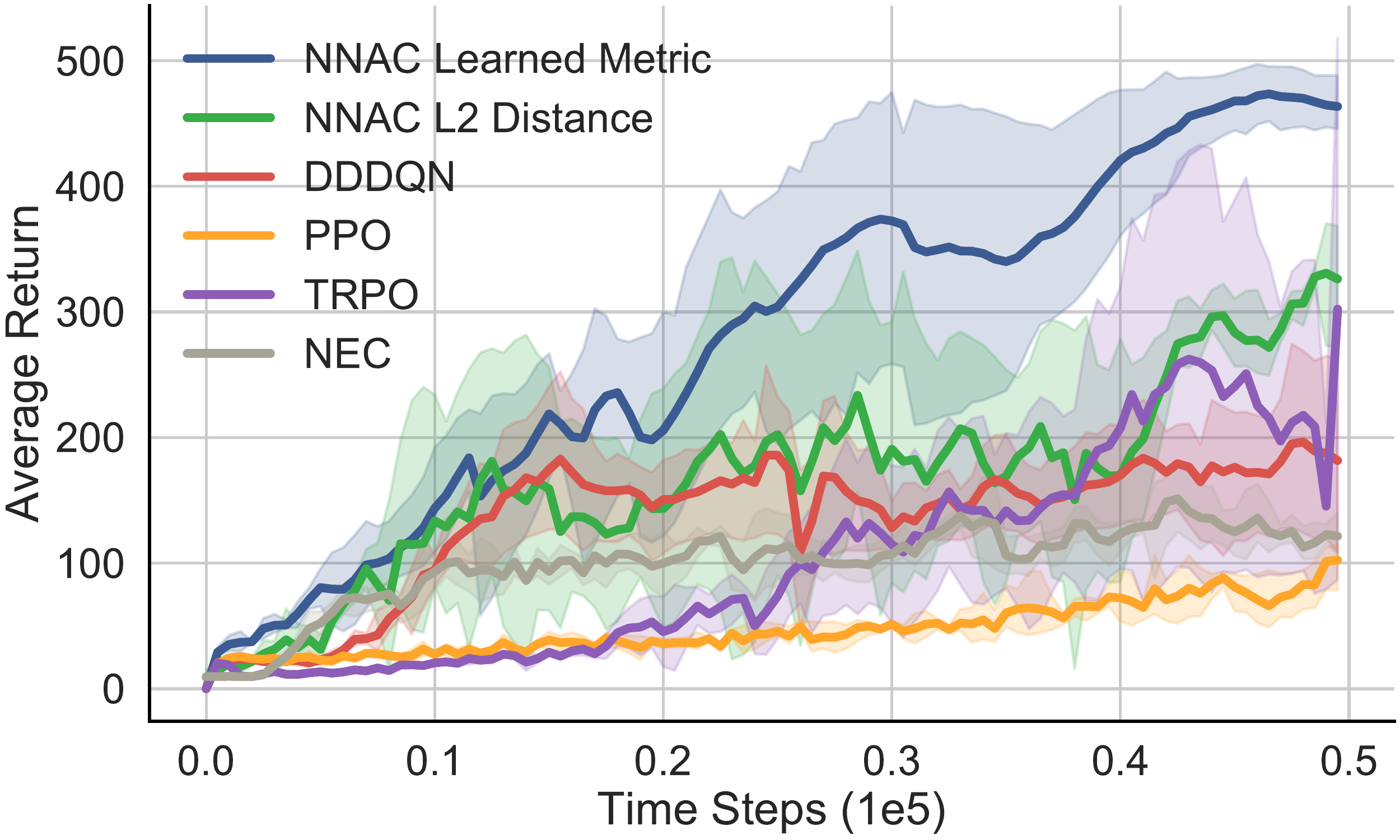}
    \caption{\small{Learning curves for image CartPole-v1. The shaded areas denote one standard deviation of evaluations over 5 trails. For visual clarity, the curves are smoothed by taking a moving average of $500$ time steps. }}
    \label{fig:cartpolepixel}
\end{figure}

We also evaluate the baseline algorithms (DDDQN, PPO, TRPO, NEC) with pixel data of the CartPole-v1 environment. The image inputs are obtained in the same way as the NNAC experiment. In particular, $dim(\hat{\mathcal{S}}) = 4 \times 20 \times 20$. For the deep RL agents, we use the Stable Baselines \cite{stable-baselines} implementation with the CNN policy. For NEC, we use the author-provided implementation. The policy networks have identical structures to the NNAC actor. Figure~\ref{fig:cartpolepixel} shows that compared with learning from internal state descriptors, learning from pixels requires more samples in general. NNAC with a learned metric solves the environment with the highest sample efficiency. Regardless of the distance metric used, NNAC obtains better average rewards at the end of training compared with the other algorithms.

\section{Example Usage of Soft NN Update Module}
In deterministic control systems, unlike neural network value function approximators which may change drastically over iterations and do not have a monotonic improvement guarantee, the nearest neighbor function approximator ensures that the value upper bound becomes smaller as the data pool size becomes larger. Approximation accuracy is improved in the sense that once a regret is paid, the algorithm can gain some new information such that the same regret will not be paid again. 

To better illustrate how the NN approximator can be combined with existing algorithms, we provide the pseudocode for two widely-used deep RL agents, DDPG \cite{Lillicrap2016ContinuousCW} and TD3 \cite{Fujimoto2018AddressingFA}, when equipped with the NN critic (Algorithm \ref{algo:NNDDPG} and \ref{algo:NNTD3}, respectively).

\begin{algorithm*}[t!]
\caption{Soft Nearest Neighbor DDPG}
\footnotesize
\begin{algorithmic}[0]
\State Randomly initialize critic network $\theta_\pi$ and value network $\theta_Q$
\State Initialize target networks $\theta'_{\pi} \leftarrow \theta_\pi$, $\theta'_{Q} \leftarrow \theta_Q$, replay buffer $B = \emptyset$, NN critic weight $\alpha \leftarrow \alpha_0$

\For{episode = $1, ..., K$}
    \State Receive initial random observation $s_1$
    \For{$h = 1, ..., H$}
        \State Take action $a_h = \pi(s_h|\theta_\pi) + N(0, \sigma)$, receive reward $r_h$ and next state $s_{h+1}$
        \State $B \leftarrow B \cup \{(s_h, a_h, s_{h+1}, r_h, h, \mathrm{NA})\}$
        
        \State Sample a mini-batch of $N$ transitions $(s_i, a_i, s_{i+1}, r_i, h_i, \delta_i)$
        \State $y_i = r_i + \gamma \cdot Q'(s_{i+1}, \pi'(s_{i+1}|\theta_{\pi'})|\theta_{Q'})$
        \State $\delta^{Q}_i = y_i - Q(s_{i}, a_i|\theta_{Q})$
        \If {$\alpha > \epsilon$}
            \State $\hat{V}(s_i) \leftarrow$ \Call{NNFuncApprox}{$s_i, h_i, \pi', B, H$}
            \State $\hat{V}(s_{i+1}) \leftarrow$ \Call{NNFuncApprox}{$s_{i+1}, h_i+1,  \pi',  B, H$} 
            \State $\delta^{NN}_i = r_i+\gamma\cdot \hat{V}(s_{i+1})- \hat{V}(s_i)$ \Comment{NN TD error estimates}
            \State Update the critic: $L'(\theta_Q) = N^{-1}\sum \alpha \left\| \delta_i^{Q} - \delta_i^{{NN}} \right\|^2 + (1 - \alpha) (y_i - Q'(s_{i},a_i|\theta_{Q'}))^2$ 
            \State Update the actor: $ \nabla_{\theta_\pi}J'(\theta_\pi) = N^{-1}\cdot \sum \alpha \cdot  \delta_i^{NN} \nabla_{\theta_\pi} \log \pi(a_i|s_i; \theta_\pi) + (1 - \alpha)\nabla_a Q(s_i,\pi(s_i|\theta_\pi)|\theta_Q) \nabla_{\theta_\pi}\pi(s_i|\theta_\pi)$ 
            \State $\delta_i \leftarrow \delta_i^{NN}$
        \Else
            \State Update the critic: $L'(\theta_Q) = N^{-1}\sum (y_i - Q'(s_{i},a_i|\theta_{Q'}))^2 + \epsilon\left\| \delta_i^{Q} - \delta_i \right\|^2$  \Comment{Continual TD supervision}
            \State Update the actor: $ \nabla_{\theta_\pi}J'(\theta_\pi) = N^{-1}\cdot \sum \nabla_a Q(s_i,\pi(s_i|\theta_\pi)|\theta_Q) \nabla_{\theta_\pi}\pi(s_i|\theta_\pi)$ 
        \EndIf
        \State $\theta'_{\pi} \leftarrow \tau \theta_\pi + (1 - \tau) \theta'_{\pi}$, $\theta'_{Q} \leftarrow \tau  \theta_Q + (1 - \tau) \theta'_Q$
    \EndFor
    \State $\alpha \leftarrow (1 - \beta)\cdot \alpha$
\EndFor
\end{algorithmic}
\label{algo:NNDDPG}
\end{algorithm*}

\begin{algorithm*}[t!]
\caption{Soft Nearest Neighbor TD3}
\footnotesize
\begin{algorithmic}[0]
\State Randomly initialize critic network $\theta_\pi$ and value networks $\theta_{Q_1}$,$\theta_{Q_2}$
\State Initialize target networks $\theta'_{\pi} \leftarrow \theta_\pi$, $\theta'_{Q} \leftarrow \theta_Q$, replay buffer $B = \emptyset$, NN critic weight $\alpha \leftarrow \alpha_0$

\For{episode $= 1,...,K$}
    \State Receive initial random observation $s_1$
    \For{$h=1,...,H$}
        \State Take action $a_h = \pi(s_h|\theta_\pi) + N(0, \sigma)$, receive reward $r_h$ and next state $s_{h+1}$
        \State $B \leftarrow B \cup \{(s_h, a_h, s_{h+1}, r_h, h, \mathrm{NA})\}$
        
        \State Sample a mini-batch of N transitions $(s_i, a_i, s_{i+1}, r_i, h_i, \delta_i)$
        \State $y_i = r_i + \gamma \cdot \min_{n=1,2} Q'(s_{i+1}, a'_{i+1}|\theta_{Q'_n})$, $a'_{i+1} \leftarrow, \pi'(s_{i+1}|\theta_{\pi'}) + \epsilon_n, \epsilon_n \leftarrow clip\big(N(0, \sigma), -c, c\big)$
        \State $\delta^{Q}_i = y_i - \min_{n=1,2}  Q'(s_{i},a_i|\theta_{Q'_n})$
        \If {$\alpha > \epsilon$}
            \State $\hat{V}(s_i) \leftarrow$ \Call{NNFuncApprox}{$s_i, h_i, \pi', B, H$}
            \State $\hat{V}(s_{i+1}) \leftarrow$ \Call{NNFuncApprox}{$s_{i+1}, h_i+1,  \pi',  B, H$} 
            \State $\delta^{NN}_i = r_i+\gamma\cdot \hat{V}(s_{i+1})- \hat{V}(s_i)$
            \State Update the critic: $L'(\theta_{Q_n}) = \argmin_{\theta_{Q_n}}N^{-1}\sum \alpha \left\| \delta_i^{Q} - \delta_i^{{NN}} \right\|^2 + (1 - \alpha) (y_i - Q_n(s_{i},a_i|\theta_{Q_n}))^2$ 
            \If{$h \mod policyfreq$}
                \State Update the actor: $ \nabla_{\theta_\pi}J'(\theta_\pi) = N^{-1}\cdot \sum \alpha \cdot  \delta_i^{NN} \nabla_{\theta_\pi} \log \pi(a_i|s_i; \theta_\pi) + (1 - \alpha)\nabla_a Q_1(s_i,\pi(s_i|\theta_\pi)|\theta_{Q_1}) \nabla_{\theta_\pi}\pi(s_i|\theta_\pi)$ 
            \EndIf
            \State $\delta_i \leftarrow \delta_i^{NN}$
        \Else
            \State Update the critic: $L'(\theta_Q) = \argmin_{\theta_{Q_i}} N^{-1}\sum (y_i - Q_i(s_{i},a_i|\theta_{Q_i}))^2+ \epsilon\left\| \delta_i^{Q} - \delta_i \right\|^2$ 
            \If{$h \mod policyfreq$}
                \State Update the actor: $ \nabla_{\theta_\pi}J'(\theta_\pi) = N^{-1}\cdot \sum \nabla_a Q_1(s_i,a_i|\theta_{Q_1}) \nabla_{\theta_\pi}\pi(s_i|\theta_\pi)$ 
            \EndIf
        \EndIf
        \State $\theta'_{\pi} \leftarrow \tau \theta_\pi + (1 - \tau) \theta'_{\pi}$, $\theta'_{Q_n} \leftarrow \tau \theta_{Q_n} + (1 - \tau) \theta'_{Q_n}, n = 1, 2$
        
    \EndFor
    \State $\alpha \leftarrow (1 - \beta)\cdot \alpha$
\EndFor

\end{algorithmic}
\label{algo:NNTD3}
\end{algorithm*}

\section{Hyperparameters of MuJoCo Experiments} \label{appendix:mujoco}
\subsection{Soft Nearest Neighbor DDPG}
\subsubsection{NN-related parameters}
\hfill
\begin{table}[ht]
    \footnotesize
    \centering
    \vspace{-5pt}
    \setlength{\tabcolsep}{3pt}
    \small{
    \begin{tabular}{ccccc}
    \toprule
    Environment & Hopper & Walker2d & HalfCheetah & Ant \\
    \midrule
    $\alpha_0$ & 0.9 & 0.5 & 0.9 & 0.9\\
     $\beta$ & 0 if $k < 20$ & 0 if $k < 20$ &0 if $k < 20$ & 0.995\\
        & 1 otherwise & 1 otherwise & 1 otherwise &\\
    $\epsilon$&  $10^{-3}$ & $10^{-3}$& $10^{-3}$ & $10^{-3}$\\
     $M$& 1 & 1 & 1 & 1 \\
     $L$ & 7 & 7 & 5 & 7 \\
     $H'$& 12 & 12 & 12 & 12\\
     $\tau_{NN}$ & 0.2 & 0.2 & 0.2 & 0.2\\
     Neg $\delta$ scale &0.3 & 0.3 & 0.3 & 0.3\\
     Grad clip& 10 & 10 & 10 & 10\\
    \bottomrule
    \end{tabular}}
    \vspace{-10pt}
\end{table}

\begin{itemize}
\small
    \item $\alpha_0$: initial NN weight.
    \item $\beta$: NN weight decreasing rate.
    \item $\epsilon$: NN termination threshold, i.e., when NN weight is smaller than $\epsilon$, we no longer use MC rollouts to estimate TD errors.
    \item $M$: number of nearest neighbor.
    \item $L$: Lipschitz parameter.
    \item $H'$: planning horizon.
    \item  $\tau_{NN}$ : target network update rate when using the NN critic.
    \item Negative TD error scale: scale the negative TD error to improve network convergence.
    \item Gradient clip threshold: clip the TD error policy gradient to improve network stability.
\end{itemize}

\subsubsection{Environment and network-related parameters}
\hfill
\begin{table}[h!]
    \footnotesize
    \centering
    \vspace{-5pt}
    \setlength{\tabcolsep}{5pt}
    \small{
    \begin{tabular}{ccccc}
    \toprule
    Environment & Hopper & Walker2d & HalfCheetah & Ant \\
    \midrule
    $\gamma$ & 0.99&0.99&0.99&0.99\\
    Net. struct. & \multicolumn{3}{c}{(400, ReLU, 300, ReLU)}\\
    Weight init. & \multicolumn{3}{c}{$[-3 \times 10^{-3}, 3 \times 10^{-3}]$}\\
    Optimizer & \multicolumn{3}{c}{Adam}\\
     $\theta_\pi$ lr &  $10^{-3}$ & $10^{-3}$ & $10^{-3}$ & $10^{-3}$\\
     $\theta_Q$ lr &  $10^{-3}$ & $10^{-3}$& $10^{-3}$ & $10^{-3}$\\
     $\sigma_N$& 0.3 & 0.2 & 0.2 & 0.1 \\
     $\tau$ &0.005&0.005&0.005&0.005\\
     Batch size & 256  & 256  & 256  & 256 \\
     $r$ scale&0.1&0.1&1&1\\
     
    \bottomrule
    \end{tabular}}
    \vspace{-10pt}
\end{table}
\begin{itemize}
\small
    \item $\gamma$: variance reduction parameter.
    \item Network structure: used in policy and value networks.
    \item Weight initialization: range of the uniform distribution for network weight initialization.
    \item Optimizer: used for both policy and value learning.
    \item $\theta_\pi$ lr: initial policy network learning rate.
    \item $\theta_Q$ lr: initial value network learning rate.
    \item $\sigma_N$: standard deviation of the normal action noise.
    \item $\tau$: target networks update parameter.
    \item Batch size: size of mini-batches at each gradient update step.
    \item $r$ scale: scale the reward to improve network stability.
\end{itemize}

\subsection{Soft Nearest Neighbor TD3}
\subsubsection{NN-related parameters}
\hfill
\begin{table}[ht]
    \footnotesize
    \centering
    \vspace{-5pt}
    \setlength{\tabcolsep}{3pt}
    \small{
    \begin{tabular}{ccccc}
    \toprule
    Environment & Hopper & Walker2d & HalfCheetah & Ant \\
    \midrule
    $\alpha_0$ & 0.9 & 0.9 & 0.9 & 0.9\\
     $\beta$ & 0 if $k < 20$ & 0 if $k < 20$ &0 if $k < 20$ & 0 if $k < 20$\\
        & 1 otherwise & 1 otherwise & 1 otherwise &1 otherwise\\
    $\epsilon$&  $10^{-3}$ & $10^{-3}$& $10^{-3}$ & $10^{-3}$\\
     $M$& 1 & 1 & 1 & 1 \\
     $L$ & 4 & 4 & 5 & 4 \\
     $H'$& 12 & 12 & 12 & 12\\
     $\tau_{NN}$ & 0.2 & 0.2 & 0.2 & 0.2\\
     Neg $\delta$ scale &0.3 & 0.3 & 0.3 & 0.3\\
     Grad clip& 10 & 10 & 10 & 10\\
    \bottomrule
    \end{tabular}}
    \vspace{-10pt}
\end{table}

\subsubsection{Environment and network-related parameters}
\hfill
\begin{table}[h!]
    \footnotesize
    \centering
    \vspace{-5pt}
    \setlength{\tabcolsep}{5pt}
    \small{
    \begin{tabular}{ccccc}
    \toprule
    Environment & Hopper & Walker2d & HalfCheetah & Ant \\
    \midrule
    $\gamma$ & 0.99&0.99&0.99&0.99\\
    Net. struct. & \multicolumn{3}{c}{(400, ReLU, 300, ReLU)}\\
    Weight init. & \multicolumn{3}{c}{$[-3 \times 10^{-3}, 3 \times 10^{-3}]$}\\
    Optimizer & \multicolumn{3}{c}{Adam}\\
     $\theta_\pi$ lr &  $10^{-3}$ & $10^{-3}$ & $10^{-3}$ & $10^{-3}$\\
     $\theta_Q$ lr &  $10^{-3}$ & $10^{-3}$& $10^{-3}$ & $10^{-3}$\\
     $\sigma_N$& 0.3 & 0.2 & 0.2 & 0.2 \\
     Noise clip &0.5&0.5&0.5&0.5\\
     $\tau$ &0.005&0.005&0.005&0.005\\
     Batch size & 256  & 256  & 256  & 256 \\
     $r$ scale&0.1&0.1&1&0.1\\
     Policy freq. &2&2&2&2\\
     
    \bottomrule
    \end{tabular}}
    \vspace{-10pt}
\end{table}
\begin{itemize}
\small
    \item Noise clip: action noise clip threshold.
    \item Policy frequency: policy update frequency w.r.t. gradient update steps.
\end{itemize}
\clearpage
\bibliography{reference}